\documentclass{article}

\usepackage{arxiv}

\usepackage{lineno,hyperref}
\usepackage{float}
\usepackage{amsmath}

\usepackage{algorithm}  
\usepackage{algorithmicx}  
\usepackage{algpseudocode}
\usepackage{enumitem}

\usepackage{multirow}
\usepackage{array} 
\newcolumntype{Y}{>{\centering\arraybackslash}X}
\usepackage{multicol}
\usepackage{makecell}

\usepackage[utf8]{inputenc} 
\usepackage[T1]{fontenc}    
\usepackage{hyperref}       
\usepackage{url}            
\usepackage{booktabs}       
\usepackage{amsfonts}       
\usepackage{nicefrac}       
\usepackage{microtype}      
\usepackage{lipsum}		
\usepackage{graphicx}
\usepackage{natbib}
\usepackage{doi}
\usepackage{subcaption}
\newcommand{\tableref}[1]{Table~{\ref{#1}}}
\newcommand{\comment}[1]{}
\usepackage{tabularx}

\title{AGA: An adaptive group alignment framework for structured medical cross-modal representation learning}

\author{Li Wei \\
	School of Computing and Artificial Intelligence\\
	Southwest Jiaotong University\\
	Chengdu, China 611756 \\
	\texttt{liweii0521@163.com} \\
	\And
	Gong Xun\thanks{corresponding author} \\
	School of Computing and Artificial Intelligence\\
	Southwest Jiaotong University\\
	Chengdu, China 611756 \\
	\texttt{xgong@home.swjtu.edu.cn} \\
	\And
	Li Jiao \\
	Department of Gastroenterology\\
	The Third People’s Hospital of Chengdu\\
	Chengdu, China 610031 \\
	\texttt{cylijiao@163.com} \\
    \And
	Sun Xiaobin \\
	Department of Gastroenterology\\
	The Third People’s Hospital of Chengdu\\
	Chengdu, China 610031 \\
	\texttt{xbsun1197@163.com}
}

\renewcommand{\shorttitle}{\textit{arXiv} Template}

\hypersetup{
pdftitle={A template for the arxiv style},
pdfsubject={q-bio.NC, q-bio.QM},
pdfauthor={David S.~Hippocampus, Elias D.~Striatum},
pdfkeywords={First keyword, Second keyword, More},
}

\begin{document}
\maketitle

\begin{abstract}
Learning medical visual representations directly from paired medical images and reports has emerged as a promising direction in representation learning. However, existing vision-language pretraining (VLP) methods in the medical domain often oversimplify clinical reports into single entities or fragmented tokens, overlooking their inherent structured nature. Moreover, contrastive learning paradigms typically rely on large quantities of hard negative samples, which poses challenges when dealing with small scale medical datasets. To address these issues, we propose Adaptive Grouped Alignment (AGA), a novel framework for learning structured information from paired medical images and reports. Specifically, we design a bidirectional grouping mechanism based on a sparse similarity matrix. Given an image-report pair, we first compute a fine-grained similarity matrix between each text token and each image patch. For each token, we select the top-matching patches to form a visual group, and conversely, for each patch, we select the most semantically related tokens to form a language group. To enable adaptive grouping, we introduce two threshold gating modules, Language-grouped Threshold Gate and Vision-grouped Threshold Gate, which dynamically learn similarity thresholds for group construction. The group representation corresponding to each token or patch is computed as a weighted average over the elements in its group, where the weights are given by their similarity scores. To align each token representation with its corresponding group representations, we propose an Instance-aware Group Alignment (IGA) loss, which operates solely within individual image-text pairs, eliminating the need for external negative samples and thereby alleviating the reliance on large scale hard negatives. Finally, we employ a Bidirectional Cross-modal Grouped Alignment (BCGA) module to facilitate fine-grained alignment between visual and linguistic group representations. Extensive experiments on both public and private datasets across various downstream tasks, including image-text retrieval and classification (in both fine-tuning and zero-shot settings), demonstrate the effectiveness of our proposed framework.
\end{abstract}


\section{Introduction}
\begin{figure}[!htb]
	\centering
		\includegraphics[width=0.55\linewidth]{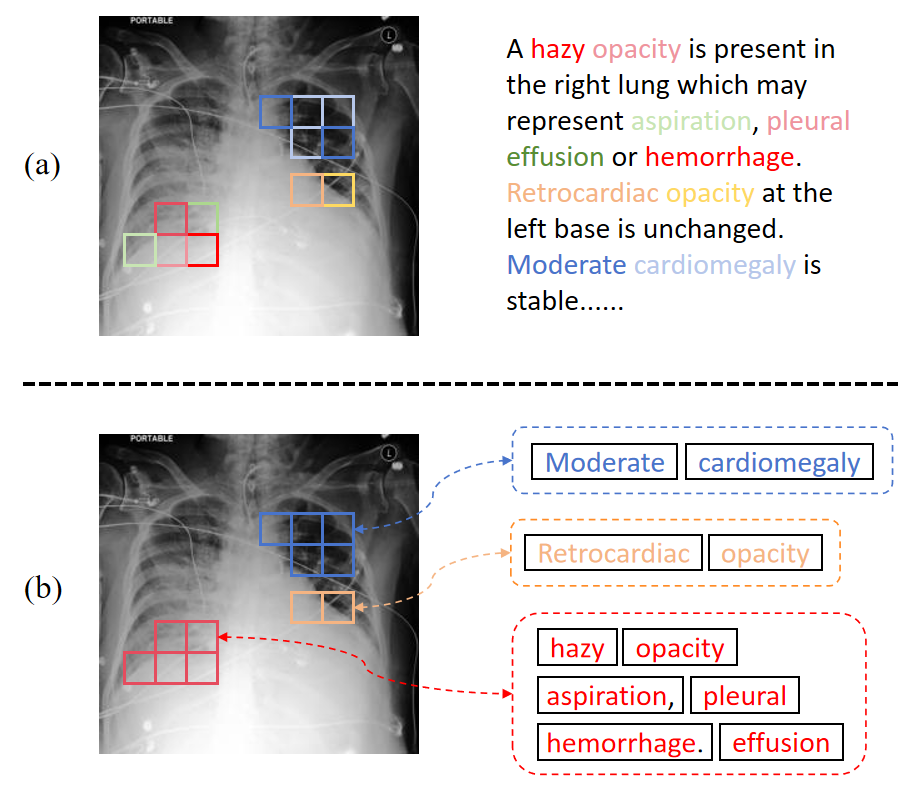}
	  \caption{Fine-grained alignment. (a) illustrates the conventional word-to-patch alignment approach. (b) shows our proposed group-wise alignment strategy, where colors denote the corresponding alignment relationships.}
    \label{fig:fig1}
\end{figure}

Advances in medical imaging technologies have revolutionized healthcare practices and significantly improved patient outcomes. However, the rapidly increasing volume of imaging studies in recent years has posed substantial challenges, including the fact that annotating medical imaging datasets requires domain expertise, imposing a growing burden on radiologists and incurring prohibitive costs at scale. Consequently, the development of effective medical image models is hindered by the scarcity of large scale manually annotated datasets. To overcome this limitation, vision-language pretraining (VLP) methods that learn visual representations of medical images directly from radiology reports without any additional manual annotation have become mainstream \cite{1,2,4}. These methods aim to learn general medical visual representations from detailed clinical narratives authored by physicians, which can then be transferred to downstream tasks. Recent works leverage such medical reports as supervisory signals and optimize the multimodal representations by maximizing the mutual information between global representations of paired images and reports \cite{4,5}. Nonetheless, considering that pathological findings often occupy only small regions within an entire image, several studies \cite{6,7} have explored learning local representations for medical language-image tasks. Furthermore, methods such as those in \cite{8,9,10} improve semantic-driven contrastive learning by segmenting medical reports into sentences instead of individual tokens, facilitating fine-grained cross-modal interactions. These advances mark significant progress in medical VLP. However, these approaches commonly oversimplify medical reports into single entities or fragmented tokens and neglect their inherent structured nature, as illustrated in Fig.~\ref{fig:fig1}(a). Sentence-level representations typically compress entire sentences into global vectors, which tend to conflate multiple independent clinical entities, anatomical locations, and attribute information. This leads to semantic ambiguity and impedes precise alignment with corresponding regions in medical images. 

Furthermore, due to the inherent difficulty in acquiring medical data, sample sizes are often limited. Traditional contrastive learning heavily relies on a large number of hard negative samples \cite{11,12, Rel, Pro}. Although prototype-based contrastive methods \cite{13,14} reduce the need for negative mining by aligning samples with class prototypes, these approaches typically depend on static class prototypes constructed across samples. Such static prototypes struggle to accommodate the presence of multiple heterogeneous semantic units within a single medical image-text pair (e.g., multiple lesions, anatomical sites, or attribute combinations). This issue is particularly pronounced under weakly supervised or unlabeled conditions, where prototypes fail to accurately capture the diversity and fine-grained semantics inherent within individual samples.

To address the aforementioned challenges, we propose a novel VLP framework with Adaptive Grouped Alignment, termed AGA. Unlike existing methods that simplify images and reports into single entities or fragmented tokens, our framework aligns local text token embeddings with their corresponding Token-Grouped Visual (TGV) embeddings using an Instance-aware Group Alignment (IGA) loss, as shown in Fig.~\ref{fig:fig1}(b). This is achieved by learning structured information among patches through a grouping mechanism. For example, a single word in a report may correspond to multiple localized visual regions. Conversely, we generate Patch-Grouped Language (PGL) embeddings by aligning each visual patch with a group of semantically related textual tokens. The proposed group representations are constructed from semantically coherent subsets of tokens or patches, enabling more natural correspondence with localized visual patterns in medical images. By associating each text token (or image patch) with its corresponding semantic group representation, the model is able to better capture hierarchical and compositional structural information, thereby facilitating more precise cross-modal alignment. To realize this, we first compute a pairwise similarity matrix for each image-text pair. To form a visual group for each text token, we introduce a Language-grouped Threshold Gate, which progressively sparsifies the similarity matrix during training by adaptively lowering the grouping threshold via a momentum-based mechanism. The resulting similarity scores are row-normalized to assign a weight to each image patch embedding, and the weighted sum of these embeddings constitutes the corresponding TGV representation, enabling dynamic grouping. A similar procedure is applied to generate the PGL embedding for each image patch by aggregating its most relevant textual tokens.

To align local token embeddings with their corresponding group representations while mitigating the reliance on a large number of hard negatives, we introduce an IGA loss. Unlike conventional prototype-based contrastive learning approaches that operate at the class level, IGA constructs multiple semantically coherent group representations within each instance, and guides each image patch or text token to align with its associated semantic group. This method not only preserves the semantic aggregation advantages of prototype-based learning, but also provides enhanced instance-level modeling capacity and structural expressiveness.

In summary, the main contributions of this work are as follows:

\begin{enumerate}
\item We propose a novel group alignment framework, termed AGA, which constructs group representations for both textual tokens and visual patches by computing a sparse similarity matrix for each image-report pair. By learning from these group-level representations, the model captures structured information and enhances cross-modal feature expressiveness.

\item To enable adaptive grouping, we introduce learnable Threshold Gates that dynamically select grouping thresholds. Additionally, we design an Instance-aware Group Alignment (IGA) loss that aligns token embeddings with their corresponding group representations. This alignment is performed within each individual image-report pair, eliminating the need for hard negatives and allowing for fine-grained representation learning in a more efficient manner.

\item We employ a Bidirectional Cross-modal Grouped Alignment (BCGA) module to align group representations across modalities, and conduct extensive experiments on both public and private medical datasets. The results on downstream tasks such as image-text retrieval and classification, under both finetuning and zero-shot settings, demonstrate the effectiveness and generalizability of the proposed framework.

\end{enumerate}  

The remainder of this paper is organized as follows. Section 2 briefly reviews related work. Section 3 presents the overall architecture of our model, including the training process and alignment process. Section 4 reports the experimental datasets and results. In Section 5, we provide a detailed discussion. Finally, Section 6 concludes the paper.

\section{Related work}
\subsection{Medical vision-language pretraining}
Vision-language pretraining (VLP) on large scale medical image-text datasets has emerged as a widely adopted paradigm for learning generic visual representations, supporting various downstream tasks and serving as a foundation for visual encoders in multimodal foundation models \cite{2, DiagLLM, 4, sparc}. By aligning global image and text representations in a shared latent space using matched and mismatched image-text pairs, these models have demonstrated strong performance in image-level vision tasks such as classification, coarse-grained retrieval, and visual question answering \cite{5, 5-1, 5-2, 5-3}. However, they often suffer from the drawback of discarding fine-grained information. To address this, Huang et al. \cite{6} made significant contributions by employing attention-based mechanisms to contrast image regions with words in paired reports, enabling the learning of localized visual representations that better capture the fine-grained patterns present in medical images. Building on this, several studies \cite{8, 9, 10, gtol} have advanced semantics-guided contrastive methods by segmenting medical reports into sentences rather than individual words, thus facilitating localized cross-modal interactions. Other works have focused on multi-scale contrastive learning. For instance, Liao et al. \cite{15} optimized the estimation of mutual information between local image features and sentence-level text representations, improving fine-grained alignment. Seibold et al. \cite{16} assumed that each sentence conveys distinct diagnostic information and proposed aligning images with corresponding sentences. Palepu and Beam \cite{17} further introduced entropy-based regularization on token representations to penalize image patch similarities, encouraging more informative alignments. Nevertheless, sentence-level embeddings compress the entire sentence into a single global vector, leading to information mixing. This makes it difficult to identify which specific words correspond to which image regions, thereby limiting the flexibility of structural modeling.

\subsection{Contrastive learning}
Contrastive learning \cite{18, 19, 20} aims to learn an embedding space in which positive instances are mapped close to each other while negative instances are pushed far apart. A key challenge in contrastive learning is the effective identification of positive and negative pairs. To improve the efficiency of contrastive learning, some studies have proposed predicting the features of one view from another view \cite{18, 21}. Furthermore, works such as  \cite{22, 23, 24, 20} have introduced the power of contrastive learning into the medical imaging domain, achieving substantial progress. Recently, several prototype-based contrastive learning methods have been proposed to leverage semantic information at the prototype level within datasets. For example, \cite{25, 26, 27} contrast instance features with their paired prototype features, while  \cite{28, 29, 7} employ clustering-based methods to perform prototype-to-prototype contrast. However, prototype alignment relies on aggregated prototypes at the class level, which is suitable for tasks with known labels and limited categories, but struggles to model fine-grained semantic structures and instance-level variations within samples.

\section{Method}

\begin{figure*}[!htb]
	\centering
	\includegraphics[width=0.9\linewidth]{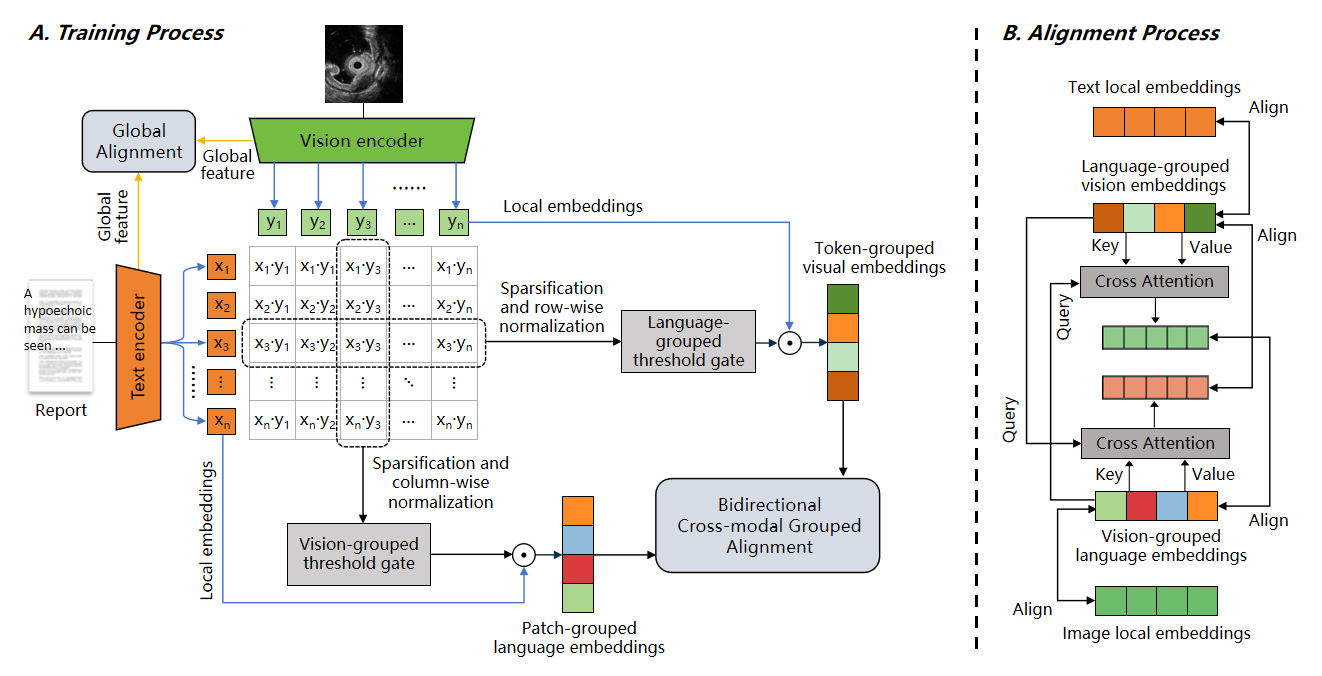}	
	\caption{. Overview of the proposed adaptive group alignment (AGA) framework. Dynamic grouping is achieved via matrix sparsification and the grouping threshold gates, while the Bidirectional Cross-modal Grouped Alignment (BCGA) module facilitates effective cross-modal grouped alignment.}
	\label{fig:fig2}
\end{figure*}

The goal of this work is to jointly learn global and local multimodal representations of medical images by leveraging medical reports, aiming to support downstream tasks with limited manual annotations. Unlike other approaches, local representations are not constructed from fragmented words or image patches but rather from novel grouped representations. Here, we first describe the image and text encoders used to extract features from each modality in Section 3.1. In Section 3.2, we formalize the computation and sparsification of the similarity matrix and introduce the threshold gates used for grouping. Finally, in Section 3.3, we present the alignment process and the associated alignment loss. The overall framework is illustrated in Fig.~\ref{fig:fig2}.

\subsection{Image and text encoding}
Given a batch of paired inputs $\left\{ {\left( {x_1^v,x_1^t} \right),...,\left( {x_b^v,x_b^t} \right)} \right\}$, where $\left( {x_i^v,x_i^t} \right)$ denotes the $i$-th image-text pair within the batch and $b$ represents the batch size, we employ an image encoder and a text encoder to extract both global and local features from each modality. These global and local features are subsequently utilized for further processing within our framework. The encoders are trained jointly with our representation learning objectives.


\textbf{Image encoding}: To construct the image encoder ${E_v}$, we adopt the ResNet-50 \cite{30} architecture as the backbone, initialized with pretrained weights from ImageNet \cite{31}. We extract local image features from the feature maps of the third bottleneck building block, while global image features are obtained from the final adaptive average pooling layer. Specifically, for the $i$-th image $x_i^v$, the corresponding patch embeddings are represented as ${V_{i,l}} = \left( {{v_{i,1}},{v_{i,2}},...,{v_{i,N}}} \right)$ with ${V_{i,N}} \in \mathbb{R} {^d}$, where $d$ denotes the feature dimension and $N$ is the number of patch embeddings. The global image embedding is denoted as ${\overline v _i} \in \mathbb{R} {^d}$.

\textbf{Text encoding}: We employ a 12-layer BioClinicalBERT \cite{32} model, pretrained on medical texts from the MIMIC-\uppercase\expandafter{\romannumeral3} dataset, as our text encoder ${E_t}$ to obtain clinically relevant text embeddings. By aggregating the embeddings from the last four layers, we derive the local (word-level) embeddings ${T_{i,l}} = \left( {{t_{i,1}},{t_{i,2}},...,{t_{i,{M_i}}}} \right)$, where each token embedding ${t_{i,{M_i}}} \in \mathbb{R} {^d}$ and ${M_i}$ denotes the number of tokens in sample $i$. The global embedding of the report is represented as ${\overline t _i} \in \mathbb{R} {^d}$. 

\subsection{Sparsification of the similarity matrix}
To learn structured information, we construct a grouped alignment by grouping words based on image patches and grouping image patches based on words. Specifically, for the $i$-th image-text pair, its similarity matrix is denoted as ${S_i} = {[{S_{i,mn}}]_{{M_i} \times N}}$, where ${s_{i,mn}} = {v_{i,n}} \cdot {t_{i,m}}$ represents the inner product between the $n$-th image patch embedding and the $m$-th word token embedding, with $n = 1,...,N$ and $m = 1,...,{M_i}$. For simplicity, we omit the index $i$ in the following discussion. To obtain alignment weights, we first apply min-max normalization along each row (i.e., for each patch) of the similarity matrix to scale values into the range $\left[ {0,1} \right]$:
\begin{equation}
\begin{array}{l}
{\widehat s_{mn}} = \frac{{{s_{mn}} - {{\min }_k}{s_{mk}}}}{{{{\max }_k}{s_{mk}} - {{\min }_k}{s_{mk}}}}
\end{array}
\end{equation}

We sparsify the similarity matrix $S = {\left( {{{\widehat s}_{jk}}} \right)_{1 \le j \le M,1 \le k \le N}}$ to facilitate learning and encourage each token to align with only a few patches, i.e.,
\begin{equation}
\begin{array}{l}
{\widetilde s_{jk}} = \left\{ {\begin{array}{*{20}{c}}
   {{{\widehat s}_{jk}},} & {if \; {{\widehat s}_{jk}} \ge \sigma }  \\
   {0,} & {otherwise}  \\
\end{array}} \right.
\end{array}
\label{eq: 2}
\end{equation}
where $\sigma$ denotes the sparsity threshold. The alignment weights are computed as:
\begin{equation}
\begin{array}{l}
{\alpha _{jk}} = \frac{{{{\widetilde s}_{jr}}}}{{\sum\nolimits_{r = 1}^R {{{\widetilde s}_{jr}}} }}
\end{array}
\end{equation}
Here, $\alpha _{jk}$ represents the weight assigned to the $k$-th visual patch embedding in the group associated with token $j$. This approach enables a flexible mapping between each token and an arbitrary number of patch embeddings in the visual domain. For each token ${t_m}$ , we compute the corresponding TGV embedding ${p_m} \in \mathbb{R} {^d}$ as:
\begin{equation}
\begin{array}{l}
{p_m} = \sum\limits_{r = 1}^R {{\alpha _{mr}}{v_r}}
\end{array}
\end{equation}
as a weighted combination of the aligned patch embeddings, where $R$ denotes the number of patches with non-zero alignment weights. 

For each image patch ${v_n}$, we obtain its corresponding PGL embedding ${q_n} \in \mathbb{R} {^d}$ in the same manner.

\subsection{Grouping threshold gate}

To sparsify the similarity matrix, we introduce a sparsity threshold $\sigma$. Unlike using a fixed threshold, we design two dynamic mechanisms: the Language-grouped Threshold Gate and the Vision-grouped Threshold Gate, which adaptively adjust the threshold during training. The dynamic threshold selection process in the Language-grouped Threshold Gate is as follows:
\begin{equation}
\begin{array}{l}
{\sigma _{tg}} = {\gamma _{tg}} \cdot {\sigma _{tg}} + \left( {1 - {\gamma _{tg}}} \right) \cdot \overline S
\end{array}
\end{equation}
where ${\gamma _{tg}}$ denotes the momentum, which controls the update rate of the threshold, and $\overline S$ represents the running average of the similarity matrix for the image-text pairs. Both image patches and text tokens originate from the same image-text pair. Patches with similarity scores higher than the historical average similarity threshold are assigned to the corresponding token groups. In contrast, patches with lower similarity scores are considered irrelevant to the group. Accordingly, Eq.~\ref{eq: 2} can be reformulated as:
\begin{equation}
\begin{array}{l}
{\widetilde s_{jk}} = \left\{ {\begin{array}{*{20}{c}}
   {{{\widehat s}_{jk}},} & {{{\widehat s}_{jk}} \ge {\sigma _{tg}}}  \\
   0, & {{{\widehat s}_{jk}} < {\sigma _{tg}}}  \\
\end{array}} \right.
\end{array}
\end{equation}

Thus, enabling dynamic threshold selection and enhancing the flexibility of grouping.

Similarly, for the Vision-grouped Threshold Gate, a dynamically updated threshold is set by operating on the transposed similarity matrix $S$:
\begin{equation}
\begin{array}{l}
{\sigma _{vg}} = {\gamma _{vg}} \cdot {\sigma _{vg}} + \left( {1 - {\gamma _{vg}}} \right) \cdot \overline {{S^T}}
\end{array}
\end{equation}
where ${\gamma _{vg}}$ represents the momentum.

\subsection{Alignment process}

\textbf{Global alignment}: To capture global information, AGA employs a global contrastive loss \cite{33,34} at the level of global image embeddings ${\overline v _i}$ and global text embeddings ${\overline t _i}$. Specifically, we optimize the similarity between each image and its corresponding text embedding, while minimizing the similarity with non-matching image-text pairs within the batch. The objective can be formulated as:
\begin{equation}
\begin{split}
\mathcal{L}_g = - \frac{1}{2b} \sum_{i=1}^{b} \bigg(
& \log \frac{\exp \left( \phi(\overline{v}_i, \overline{t}_i)/\tau_1 \right)}
{\sum_{j=1}^{b} \exp \left( \phi(\overline{v}_i, \overline{t}_j)/\tau_1 \right)} \\
& + \log \frac{\exp \left( \phi(\overline{t}_i, \overline{v}_i)/\tau_1 \right)}
{\sum_{j=1}^{b} \exp \left( \phi(\overline{t}_i, \overline{v}_j)/\tau_1 \right)} \bigg)
\end{split}
\end{equation}
where $\phi \left( {{{\overline v }_i},{{\overline t }_j}} \right) = \frac{{{{\overline v }_i}}}{{{{\left\| {{{\overline v }_i}} \right\|}_2}}} \cdot \frac{{{{\overline t }_j}}}{{{{\left\| {{{\overline t }_j}} \right\|}_2}}}$, and $\tau$ denotes the temperature parameter.

\textbf{Grouping alignment}: The group alignment process, as illustrated in Fig.~\ref{fig:fig2}b, consists of two components. For the $i$-th image-text pair $\left( {x_i^v,x_i^t} \right)$, we first perform fine-grained alignment between text token embeddings and their corresponding TGV embeddings, as well as between image patch embeddings and their corresponding PGL embeddings. We introduce the IGA loss, which operates at the level of individual image-text pairs over sequences of tokens and patches, without requiring other pairs as negative samples. This design significantly reduces both computational and memory overhead. The IGA losses for text tokens and image patches are denoted as ${L_{tf}}$ and ${L_{vf}}$, respectively:
\begin{equation}
\begin{split}
\mathcal{L}_{tf} = - \frac{1}{2b} \sum_{i=1}^{b} \Bigg[
\frac{1}{M_i} \sum_{j=1}^{M_i} \bigg(
& \log \frac{\exp\left( \phi(p_i^j, t_i^j)/\tau_2 \right)}
{\sum\limits_{k=1}^{M_i} \exp\left( \phi(p_i^j, t_i^k)/\tau_2 \right)} \\
& + \log \frac{\exp\left( \phi(t_i^j, p_i^j)/\tau_2 \right)}
{\sum\limits_{k=1}^{M_i} \exp\left( \phi(t_i^j, p_i^k)/\tau_2 \right)}
\bigg) \Bigg]
\end{split}
\end{equation}

\begin{equation}
\begin{split}
\mathcal{L}_{vf} = - \frac{1}{2b} \sum_{i=1}^{b} \Bigg[
\frac{1}{N} \sum_{j=1}^{N} \bigg(
& \log \frac{\exp\left( \phi(q_i^j, v_i^j)/\tau_2 \right)}
{\sum\limits_{k=1}^{N} \exp\left( \phi(q_i^j, v_i^k)/\tau_2 \right)} \\
& + \log \frac{\exp\left( \phi(v_i^j, q_i^j)/\tau_2 \right)}
{\sum\limits_{k=1}^{N} \exp\left( \phi(v_i^j, q_i^k)/\tau_2 \right)}
\bigg) \Bigg]
\end{split}
\end{equation}
Here, $t_i^j$ and $v_i^j$ denote the $j$-th text token embedding and image patch embedding for the $i$-th image-text pair, respectively. $p_i^j$ and $q_i^j$ represent the $j$-th TGV and PGL embeddings, respectively. The IGA loss aims to maximize the similarity between each token or patch embedding and its corresponding TGV or PGL embedding, while minimizing its similarity with other TGV or PGL embeddings within the same sequence, and vice versa.

Secondly, we perform grouped alignment between the TGV embeddings and PGL embeddings. To explicitly match and align cross-modal grouped representations between medical images and radiology reports, we adopt an efficient Bidirectional Cross-modal Grouped Alignment (BCGA) module. This module employs a cross-attention mechanism \cite{35} to compute soft alignments between the generated TGV and PGL embeddings. Specifically, for the $i$-th image-text pair, the generated TGV embeddings and PGL embeddings are first normalized, resulting in ${P_i} = \left\{ {p_i^1,p_i^2,...,p_i^{{M_i}}} \right\}$ and ${Q_i} = \left\{ {q_i^1,q_i^2,...,q_i^N} \right\}$, where ${p_i} \in \mathbb{R} {^d}$, ${q_i} \in \mathbb{R} {^d}$. For each TGV embedding $p_i^j$, we attend over all PGL embeddings ${Q_i}$, and compute the corresponding cross-modal TGV embedding $u_i^j$,
\begin{equation}
\begin{array}{l}
\begin{array}{*{20}{c}}
   {u_i^j = \sum\limits_{k = 1}^N {o\left( {\beta _i^{j2k}\left( {Vq_i^k} \right)} \right),} }  \\
   {\beta _i^{j2k} = soft\max \left( {\frac{{{{\left( {Qp_i^j} \right)}^T}\left( {Kq_i^k} \right)}}{{\sqrt d }}} \right)}  \\
\end{array}
\end{array}
\end{equation}
Here, $Q \in \mathbb{R} {^{d \times d}}$, $K \in \mathbb{R} {^{d \times d}}$ and $V \in \mathbb{R} {^{d \times d}}$ denote learnable projection matrices. After that, we apply the grouped language-to-vision alignment loss ${L_{gla}}$ , which encourages each TGV embedding $p_i^j$ to be close to its corresponding cross-modal TGV embedding $u_i^j$, while pushing it away from other cross-modal TGV embeddings. This objective effectively maximizes a lower bound on the group-level cross-modal mutual information within each image-report pair \cite{36}. The ${L_{gla}}$ loss is formulated as follows:
\begin{equation}
\begin{split}
\mathcal{L}_{gla} = - \frac{1}{2b} \sum_{i=1}^{b} \Bigg[
\frac{1}{M_i} \sum_{j=1}^{M_i} \bigg(
& \log \frac{\exp\left( \phi(p_i^j, u_i^j)/\tau_3 \right)}
{\sum\limits_{k=1}^{M_i} \exp\left( \phi(p_i^j, u_i^k)/\tau_3 \right)} \\
& + \log \frac{\exp\left( \phi(u_i^j, p_i^j)/\tau_3 \right)}
{\sum\limits_{k=1}^{M_i} \exp\left( \phi(u_i^j, p_i^k)/\tau_3 \right)}
\bigg) \Bigg]
\end{split}
\end{equation}

Similarly, we obtain the cross-modal PGL embeddings $w_i^j$, and define the grouped vision-to-language alignment loss ${L_{gva}}$ as follows:
\begin{equation}
\begin{split}
\mathcal{L}_{gva} = - \frac{1}{2b} \sum_{i=1}^{b} \Bigg[
\frac{1}{N} \sum_{j=1}^{N} \bigg(
& \log \frac{\exp\left( \phi(q_i^j, w_i^j)/\tau_3 \right)}
{\sum\limits_{k=1}^{N} \exp\left( \phi(q_i^j, w_i^k)/\tau_3 \right)} \\
& + \log \frac{\exp\left( \phi(w_i^j, q_i^j)/\tau_3 \right)}
{\sum\limits_{k=1}^{N} \exp\left( \phi(w_i^j, q_i^k)/\tau_3 \right)}
\bigg) \Bigg]
\end{split}
\end{equation}

The final loss function is defined as follows:
\begin{equation}
\begin{array}{l}
{\mathcal{L}_{total}} = {\lambda _1}{\mathcal{L}_g} + \frac{{{\lambda _2}}}{2}\left( {{\mathcal{L}_{tf}} + {\mathcal{L}_{vf}}} \right) + \frac{{{\lambda _3}}}{2}\left( {{\mathcal{L}_{gla}} + {\mathcal{L}_{gva}}} \right)
\end{array}
\end{equation}
where ${\lambda _1}$, ${\lambda _2}$, and ${\lambda _3}$ are hyperparameters that balance different components of the alignment process.

\section{Experiments}

We first introduce the paired dataset used for contrastive pretraining, the datasets employed for downstream task evaluation, and the baseline methods for comparison.

\subsection{Experimental datasets}
\subsubsection{Datasets for Pretraining}

\textbf{MIMIC-CXR} \cite{37}: We utilize the second version of the publicly available MIMIC-CXR dataset, a large scale and openly accessible collection of chest radiographs paired with corresponding radiology reports. Consistent with prior work \cite{38, 39, 40}, we use the \textit{findings} section of the raw radiology reports as reference texts. After preprocessing, the training, validation, and test sets contain 270,742 (152142), 2130 (1196), and 3858 (2347) image-report pairs, respectively. The numbers in parentheses indicate splits based on the unique ``study\_id''.

\textbf{SMTs}: We collected a private dataset consisting of image-report pairs related to submucosal tumors (SMTs) of the gastrointestinal tract from the Third People's Hospital of Chengdu. The study was conducted in accordance with the Declaration of Helsinki and was approved by the Ethics Committee of the Third People's Hospital of Chengdu on September 25, 2024 (IRB No. 2023-S-48-1). Due to the retrospective nature of the study, the requirement for informed consent was waived. To ensure patient privacy, all personally identifiable information has been removed. The dataset includes EUS (Endoscopic Ultrasound) images and corresponding textual reports collected from five different hospitals. We used data from four hospitals to construct our pretraining dataset, comprising 2455 (547), 266 (68), and 600 (120) EUS image-report pairs for the training, validation, and test sets, respectively. The numbers in parentheses indicate the quantities grouped by patient ID. Data from the same patient were assigned exclusively to a single split to ensure fair and non-overlapping partitioning. The data from the remaining hospital were reserved for downstream task evaluation.

\subsubsection{Downstream Task Datasets}

\textbf{CheXpert 5×200}: The dataset contains five abnormality categories sampled from the CheXpert dataset \cite{41}, namely Atelectasis, Cardiomegaly, Consolidation, Edema, and Pleural Effusion, with 200 image-report pairs per category \cite{6, 4}. Each instance in the dataset corresponds to a single abnormality category.

\textbf{RSNA Pneumonia} \cite{42}: We use the training set from the second phase version of this dataset, as test set labels are not available. This subset contains approximately 26700 frontal chest X-rays. The task is to classify whether each chest image exhibits Lung Opacity. 

\textbf{SMTs 3×200}: We use the SMTs pretraining test set as a downstream task subset, which includes three tumor categories (gastrointestinal stromal tumors (GISTs), neuroendocrine tumors (NETs) and Leiomyomas) randomly sampled from four hospitals, with 200 image-text pairs per category. Each instance in the dataset belongs to a single tumor category.

\textbf{SMTs SN}: This dataset is an image-text dataset with the same structure as the SMTs dataset, obtained from a single hospital. It contains 584 pairs of data from 120 patients, covering four categories: GISTs, NETs, leiomyomas, and others. Each example belongs to a single abnormality category.

\subsection{Experimental results}
\subsubsection{Implementation details}
The experiments are conducted on a platform with four NVIDIA GeForce RTX 3090 GPUs running Ubuntu 20.04, using Python 3.7.16 and PyTorch 1.12.1. The maximum number of training epochs is set to 50, with a batch size of 48. The optimizer used is AdamW, with an initial learning rate of $5 \times {10^{ - 5}}$ for both the SMTs and MIMIC-CXR datasets. For the SMTs and MIMIC datasets, the momentum hyperparameters ${\gamma _{tg}}$ (or ${\gamma _{vg}}$) are set to 0.99 and 0.999, respectively. Following the practice in contrastive learning \cite{4,24}, the embedding dimension $d$ is set to 128, and the temperature hyperparameters ${\tau _1}$, ${\tau _2}$ and ${\tau _3}$ are set to 0.3, 0.3, and 0.1, respectively. The loss weights ${\lambda _1}$, ${\lambda _2}$ and ${\lambda _3}$ are all set to 0.5 by default. For comparison, we select the classical methods ConVIRT \cite{4}, Gloria \cite{6}, MGCA \cite{7}, and SPARC \cite{sparc}. These methods are reproduced on the same datasets using identical image and text encoders to ensure a fair comparison of different alignment strategies. We pretrain these models separately on the two pretraining datasets and apply them to their corresponding downstream tasks for evaluation.

\subsubsection{Image-text retrieval}
We first evaluate the effectiveness of our representation learning framework on image-text retrieval using three datasets: CheXpert 5×200, SMTs 3×200, and SMTs SN. Following the setup in \cite{6}, given an image as the input query, the goal is to retrieve the target report by computing similarity scores between the query image and all candidate reports using the learned representations. By checking whether the retrieved report belongs to the same category as the query image, we evaluate retrieval accuracy using the Precision@$K$ metric. The top-$K$ precision scores are reported for $K$ = 5, 10, and 100.

\tableref{tab:1} presents the image-to-text retrieval results on the downstream CheXpert 5×200 dataset after pretraining the models on the MIMIC-CXR dataset. As shown in the table, our proposed AGA framework consistently outperforms other baselines. Compared to methods incorporating local contrastive losses, such as Gloria \cite{6}, MGCA \cite{7}, and SPARC \cite{sparc}, AGA achieves superior performance, reaching a Precision@5 of 50.28. This demonstrates that the group alignment strategy provides more effective structured feature representations than conventional fine-grained alignment approaches. \tableref{tab:2} reports the image-to-text retrieval results on the downstream SMTs 3×200 and SMTs SN datasets, following pretraining on the SMTs dataset. Our model achieves Precision@5 scores of 55 and 42.43, respectively. These results further highlight the competitiveness of the proposed model in capturing discriminative cross-modal representations under varying clinical data distributions.

\begin{table}[htbp]
\renewcommand\arraystretch{1}
\centering
\small
\caption{Results of image-to-text retrieval on the CheXpert 5×200 dataset.}
\begin{tabular}{l|c|c|c}
\hline
\multicolumn{1}{c|}{\multirow{2}{*}{Method}} & \multicolumn{3}{c}{CheXpert 5×200}               \\ \cline{2-4} 
\multicolumn{1}{c|}{}                        & Prec@5         & Prec@10        & Prec@100       \\ \hline
ConVIRT\cite{4}                              & 48.30          & 47.92          & 42.00          \\
Gloria\cite{6}                               & 45.56          & 43.78          & 39.26          \\
MGCA\cite{7}                                 & 49.54          & 49.10          & 42.25          \\
SPARC\cite{sparc}                            & 47.44          & 47.36          & 42.73          \\
AGA                                          & \textbf{50.28} & \textbf{49.84} & \textbf{43.80} \\ \hline
\end{tabular}
\label{tab:1}
\end{table}

\begin{table*}[]
\renewcommand\arraystretch{1}
\centering
\small
\caption{Results of image-to-text retrieval on the SMTs 3×200 and SMTs SN datasets.}
\begin{tabular}{l|ccc|ccc}
\hline
\multicolumn{1}{c|}{\multirow{2}{*}{Method}} & \multicolumn{3}{c|}{SMTs 3×200}               & \multicolumn{3}{c}{SMTs SN}                      \\ \cline{2-7} 
\multicolumn{1}{c|}{}                        & Prec@5      & Prec@10        & Prec@100       & Prec@5         & Prec@10        & Prec@100       \\ \hline
ConVIRT\cite{4}                              & 48.83       & 48.67          & 43.14          & 40.68          & 39.35          & 37.23          \\
Gloria\cite{6}                               & 54.17       & 52.33          & 42.99          & 34.93          & 34.93          & \textbf{38.60} \\
MGCA\cite{7}                                 & 48.83       & 50.83          & \textbf{44.58} & 36.44          & 36.03          & 38.06          \\
SPARC\cite{sparc}                            & 44.33       & 45.00          & 41.40          & 27.40          & 32.81          & 32.35          \\
AGA                                          & \textbf{55.00} & \textbf{54.42} & 43.16          & \textbf{42.43} & \textbf{45.07} & 37.40           \\ \hline
\end{tabular}
\label{tab:2}
\end{table*}

\subsubsection{Image Classification}
We further evaluate the learned feature representations on two distinct image classification tasks: supervised classification and zero-shot classification. 

\textbf{Supervised Classification.} For supervised classification, following the setting in \cite{4}, we attach a linear layer to the pretrained image encoder and train the model using varying proportions of labeled data (1\%, 10\%, and 100\%) to evaluate the data efficiency of global image representations. \tableref{tab:3} reports the area under the ROC curve (AUC) on the downstream CheXpert 5×200 and RSNA Pneumonia datasets. As the amount of training data increases, the AUC also improves. When fine-tuned with only 1\% of the training data, our model achieves AUC scores of 56.1 and 68.91 on CheXpert 5×200 and RSNA Pneumonia, respectively, both surpassing the baseline methods and demonstrating superior data efficiency. Due to the limited sample size in the private SMTs 3×200 and SMTs SN datasets, which is insufficient to support a 1\% fine-tuning scenario, we merge them into a unified dataset, SMTs 3×200-SN, for the supervised classification task. \tableref{tab:4} shows the results on the SMTs 3×200-SN dataset, where we observe a similar trend as in \tableref{tab:3}. When fine-tuned with 10\%, 50\%, and 100\% of the training data, our model achieves AUC scores of 57.04, 59.28, and 83.71, respectively, consistently outperforming baseline methods.

\begin{table}[]
\renewcommand\arraystretch{1}
\centering
\small
\caption{Results of linear classification on CheXpert 5×200 and RSNA with 1\%, 10\%, 100\% training data.}
\begin{tabular}{l|ccc|ccc}
\hline
\multicolumn{1}{c|}{\multirow{2}{*}{Method}} & \multicolumn{3}{c|}{CheXpert 5×200 (AUC)}       & \multicolumn{3}{c}{RSNA Pneumonia (AUC)}         \\ \cline{2-7} 
\multicolumn{1}{c|}{}                        & 1\%           & 10\%           & 100\%          & 1\%            & 10\%           & 100\%          \\ \hline
ConVIRT\cite{4}                              & 46.62         & 56.67          & 83.69          & 66.20          & 73.08          & 86.40          \\
Gloria\cite{6}                               & 54.75         & 57.81          & 84.20          & 67.84          & 74.51          & 86.92          \\
MGCA\cite{7}                                 & 55.21         & 59.21          & 83.92          & 67.28          & \textbf{76.12} & 87.76          \\
SPARC\cite{sparc}                            & 54.12         & 58.12          & 83.65          & 68.21          & 75.23          & 87.47          \\
AGA                                          & \textbf{56.10} & \textbf{61.31} & \textbf{84.32} & \textbf{68.91} & 75.78          & \textbf{87.92} \\ \hline
\end{tabular}
\label{tab:3}
\end{table}

\begin{table}[]
\renewcommand\arraystretch{1}
\centering
\small
\caption{Results of linear classification on SMTs 3×200-SN with1\%, 10\%, 100\% training data.}
\begin{tabular}{l|c|c|c}
\hline
\multicolumn{1}{c|}{\multirow{2}{*}{Method}} & \multicolumn{3}{c}{SMTs 3×200-SN(AUC)}           \\ \cline{2-4} 
\multicolumn{1}{c|}{}                        & 10\%           & 50\%           & 100\%          \\ \hline
ConVIRT\cite{4}                               & 49.91          & 54.96          & 83.63          \\
Gloria\cite{6}                                & 50.75          & 59.17          & 80.78          \\
MGCA\cite{7}                                  & 55.67          & 55.51          & 83.00          \\
SPARC\cite{sparc}                             & 45.94          & 49.59          & 80.55          \\
AGA                                          & \textbf{57.04} & \textbf{59.28} & \textbf{83.71} \\ \hline
\end{tabular}
\label{tab:4}
\end{table}

\textbf{Zero-shot classification.} For zero-shot classification, we use an image as input with the objective of predicting the corresponding label, even though the model is not explicitly trained with class labels. Inspired by \cite{6}, we convert each classification category into a textual prompt. Specifically, for datasets involving chest diseases, we adopt the prompt engineering strategy from \cite{6} to generate representative textual descriptions for each category, capturing possible subtypes, severity levels, and anatomical locations of the medical conditions. Subsequently, all category prompts and the input image are projected into a shared multimodal embedding space using the pretrained representation learning model, and the label associated with the prompt that has the highest similarity score is selected as the prediction.

\tableref{tab:5} reports the zero-shot classification results on the CheXpert 5×200 and RSNA Pneumonia datasets, with our method achieving accuracies of 63.6 and 51.1, respectively, outperforming the baseline models. Compared to existing fine-grained alignment approaches such as GLoRIA and MGCA, our group-wise alignment strategy captures fine-grained semantic relationships between image regions and textual descriptions more effectively. \tableref{tab:6} presents the results on the private SMTs 3×200 and SMTs SN datasets, where our model achieves accuracies of 56.5 and 61.5, respectively, further demonstrating its effectiveness.

\begin{table}[]
\renewcommand\arraystretch{1}
\centering
\caption{Results of zero-shot image classification on the CheXpert 5×200 and RSNA datasets.}
\begin{tabular}{l|ccc|ccc}
\hline
\multicolumn{1}{c|}{\multirow{2}{*}{Method}} & \multicolumn{3}{c|}{CheXpert 5×200}       & \multicolumn{3}{c}{RSNA Pneumonia}            \\ \cline{2-7} 
\multicolumn{1}{c|}{}                        & ACC           & ${F_1}$          & ROC         & ACC           & ${F_1}$            & ROC           \\ \hline
ConVIRT\cite{4}                              & 63.4          & 25.9        & \textbf{53.0} & 50.3          & 36.5          & 53.5          \\
Gloria\cite{6}                               & 62.4          & 22.9        & 50.8        & 50.3          & 36.0            & 54.7             \\
MGCA\cite{7}                                 & 62.3          & 25.1        & 50.3        & 50.2          & \textbf{40.9} & 54.9          \\
SPARC\cite{sparc}                            & 61.0          & 23.6        & 48.8        & 49.8          & 38.6          & 54.5          \\
AGA                                          & \textbf{63.6} & \textbf{26.0} & 52.4        & \textbf{51.1} & 37.6          & \textbf{55.3} \\ \hline
\end{tabular}
\label{tab:5}
\end{table}

\begin{table}[]
\renewcommand\arraystretch{1}
\centering
\caption{Results of zero-shot image classification on the SMTs 3×200 and SMTs SN datasets.}
\begin{tabular}{l|ccc|ccc}
\hline
\multicolumn{1}{c|}{\multirow{2}{*}{Method}} & \multicolumn{3}{c|}{SMTs 3×200}               & \multicolumn{3}{c}{SMTs SN}                   \\ \cline{2-7} 
\multicolumn{1}{c|}{}                        & ACC           & ${F_1}$            & ROC           & ACC           & ${F_1}$            & ROC           \\ \hline
ConVIRT\cite{4}                              & 55.2          & 29.5          & 44.9          & 61.0          & 27.5          & 49.5          \\
Gloria\cite{6}                               & 56.4          & 30.4          & 48.6          & 61.3          & 25.8          & 46.2          \\
MGCA\cite{7}                                 & 55.6          & \textbf{32.7} & 42.8          & 61.0          & \textbf{28.2} & 53.6          \\
SPARC\cite{sparc}                            & 53.0          & 30.9          & 46.0          & 58.5          & 25.8          & 45.9          \\
AGA                                          & \textbf{56.5} & 28.9          & \textbf{49.3} & \textbf{61.5} & 24.2          & \textbf{55.4} \\ \hline
\end{tabular}
\label{tab:6}
\end{table}

\subsubsection{Ablation studies}
This section evaluates the contribution of different components in our method. We conduct ablation studies on three variants of the pretraining setup: (1) Only Global Alignment, (2) Removing the BCGA module, and (3) Using a Fixed Threshold. These experiments are designed to assess the effectiveness of our group alignment strategy, the IGA Loss, and the dynamic threshold gating mechanism. For the fixed threshold setting, we set the parameters to ${\sigma _{tg}}$ = 1/361 and ${\sigma _{vg}}$ = 1/97, where 361 corresponds to the default number of image patch embeddings and 97 is the default maximum number of text tokens. This ensures that each text token and image patch receives a corresponding group representation. 

\tableref{tab:7} presents the image-to-text retrieval results on the CheXpert 5×200 dataset under different pretraining settings. Compared to our full model, the Only Global Alignment setting shows a performance drop of approximately 4\%, indicating that our group alignment strategy is effective in capturing fine-grained semantics beyond global features. Notably, the Removing the BCGA module setting yields the largest performance degradation, highlighting the importance of inter-group alignment. Without the BCGA module, the model relies solely on IGA to perform intra-group alignment between text tokens and its TGV embeddings. However, the absence of inter-group alignment leads to semantically disjoint group representations, which hinders the model’s ability to reason over the global context. Additionally, the groups constructed through IGA may contain noisy or ambiguous associations that cannot be refined without BCGA. As a result, the model's performance significantly degrades, even falling below the level of only global alignment, which emphasizes the critical role of the joint use of BCGA and IGA in achieving fine-grained semantic integration and contextual consistency. The Fixed Threshold variant shows the smallest performance drop (approximately 3\%), which first demonstrates the overall robustness of our framework in capturing fine-grained information. Moreover, the adaptive threshold gating module further improves overall performance by dynamically adjusting the grouping thresholds, validating the effectiveness of the adaptive mechanism. \tableref{tab:8} reports the image-to-text retrieval performance on the SMTs 3×200 and SMTs SN datasets under the same settings. Similar trends are observed, which further supports the consistency and generalization of our findings.

\begin{table}[]
\renewcommand\arraystretch{1}
\centering
\small
\caption{Ablation results on the image-to-text retrieval task on the CheXpert 5×200 dataset.}
\begin{tabular}{l|ccc}
\hline
\multicolumn{1}{c|}{\multirow{2}{*}{Method}} & \multicolumn{3}{c}{CheXpert 5×200}                                                      \\ \cline{2-4} 
\multicolumn{1}{c|}{}                        & \multicolumn{1}{l}{Prec@5} & \multicolumn{1}{l}{Prec@10} & \multicolumn{1}{l}{Prec@100} \\ \hline
Only global alignment                        & 45.34                      & 45.04                       & 40.96                        \\
No BCGA                                      & 34.28                      & 33.83                       & 31.21                        \\
AGA(fixed)                                   & 48.54                      & 47.12                       & 39.14                        \\
AGA                                          & \textbf{50.28}             & \textbf{49.84}              & \textbf{43.80}                \\ \hline
\end{tabular}
\label{tab:7}
\end{table}

\begin{table*}[]
\renewcommand\arraystretch{1}
\centering
\small
\caption{Ablation results on the image-to-text retrieval task on the SMTs 3×200 and SMTs SN datasets.}
\begin{tabular}{l|ccc|ccc}
\hline
\multicolumn{1}{c|}{\multirow{2}{*}{Method}} & \multicolumn{3}{c|}{SMTs 3×200}               & \multicolumn{3}{c}{SMTs SN}                      \\ \cline{2-7} 
\multicolumn{1}{c|}{}                        & Prec@5      & Prec@10        & Prec@100       & Prec@5         & Prec@10        & Prec@100       \\ \hline
Only global alignment                        & 48.83       & 48.67          & 43.14          & 40.68          & 39.35          & 37.23          \\
No BCGA                                      & 45.33       & 44.58          & 42.15          & 39.55          & 37.84          & \textbf{38.09} \\
AGA(fixed)                                   & 51.00       & 52.42          & 42.41          & 38.84          & 38.49          & 36.54          \\
AGA                                          & \textbf{55.00} & \textbf{54.42} & \textbf{43.16} & \textbf{42.43} & \textbf{45.07} & 37.40        \\ \hline
\end{tabular}
\label{tab:8}
\end{table*}

\section{Discussion}
\subsection{Variation of Grouping Thresholds}

\begin{figure}[!htb]
	\centering
		\includegraphics[width=0.5\linewidth]{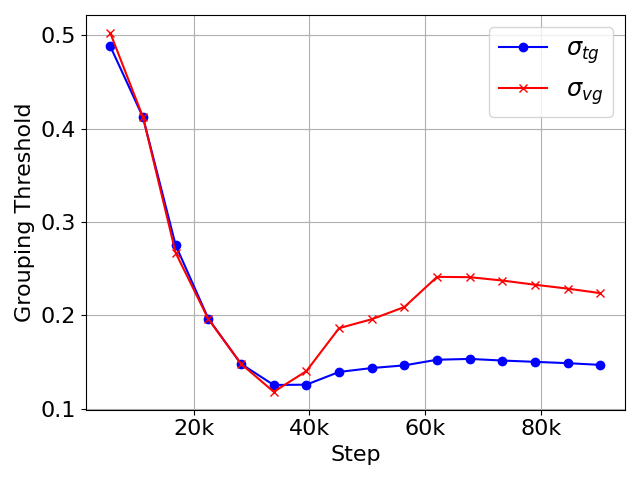}
	  \caption{Line plot of language grouping threshold ${\sigma _{tg}}$ and visual grouping threshold ${\sigma _{vg}}$ on the MIMIC-CXR dataset during pretraining.}
    \label{fig:fig3}
\end{figure}

We visualize the variation of grouping thresholds during model pretraining using line plots. As shown in Fig.~\ref{fig:fig3}, the curves illustrate the evolution of the language grouping threshold ${\sigma _{tg}}$ and the visual grouping threshold ${\sigma _{vg}}$ over optimization steps on the MIMIC-CXR dataset. It can be observed that ${\sigma _{tg}}$ stabilizes around 0.15, while ${\sigma _{vg}}$ stabilizes around 0.22, with a noticeable upward trend in the later stages. The fact that ${\sigma _{vg}}$ surpasses ${\sigma _{tg}}$ indicates that each image patch in a chest X-ray tends to align with an declining number of textual tokens. One possible explanation is that in the MIMIC-CXR dataset, findings related to specific anatomical regions are often described across multiple sentences. The language tends to be more loosely structured and unstructured. For example, the statements “There is a right lower lobe opacity.” and “This may represent pneumonia.” are independent sentences, yet together they describe the same abnormality in the right lower lobe.

\begin{figure}[!htb]
	\centering
		\includegraphics[width=0.5\linewidth]{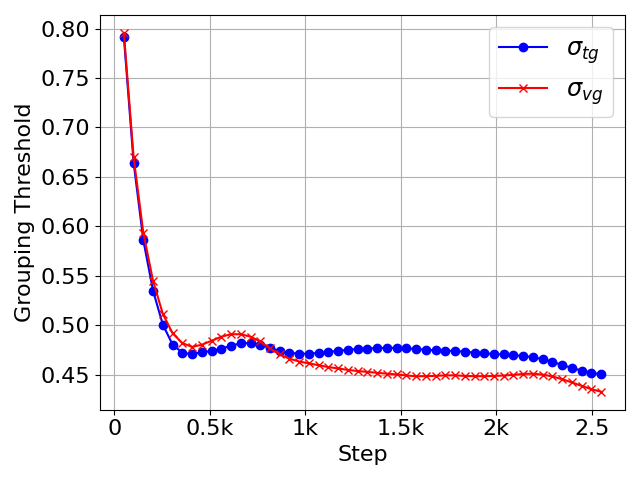}
	  \caption{Line plot of language grouping threshold ${\sigma _{tg}}$ and visual grouping threshold ${\sigma _{vg}}$ on the SMTs dataset during pretraining.}
    \label{fig:fig4}
\end{figure}

In contrast, Fig.~\ref{fig:fig4} also presents the line plots of the thresholds ${\sigma _{tg}}$ and ${\sigma _{vg}}$ over optimization steps on the SMTs dataset. It can be observed that ${\sigma _{tg}}$ stabilizes around 0.45 and ${\sigma _{vg}}$ around 0.43, with both values being nearly equal. This indicates a strong correlation between individual textual tokens and single image patches. This phenomenon can be interpreted in the dataset context as the private SMTs dataset typically features sentences that comprehensively describe multiple attributes of a single region. The descriptions are more structured and focused, for example: “A hypoechoic mass can be seen in the lesion, which is oval in shape, protruding into and outside the cavity, with...”. Coreference is rare, and the association between words and image regions is strong, which contrasts with the more loosely structured and distributed descriptions seen in the MIMIC-CXR dataset.

\subsection{Visualization of attention weights}

\begin{figure}[htbp]
	\centering
	\begin{minipage}{0.15\linewidth}
		\centering
		\includegraphics[width=\linewidth]{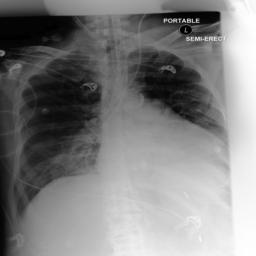}
        \includegraphics[width=\linewidth]{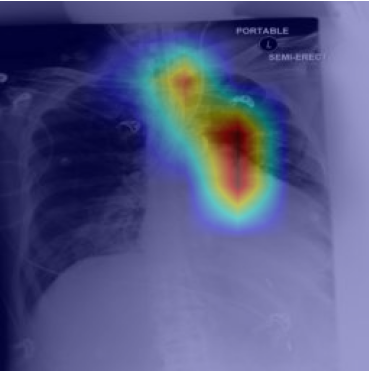}
        \parbox[c][3.5em][t]{\linewidth}{\centering \small Atelectasis}
	\end{minipage}
    \hspace{0.01\linewidth}
	\begin{minipage}{0.15\linewidth}
		\centering
		\includegraphics[width=\linewidth]{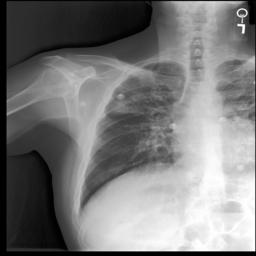}
        \includegraphics[width=\linewidth]{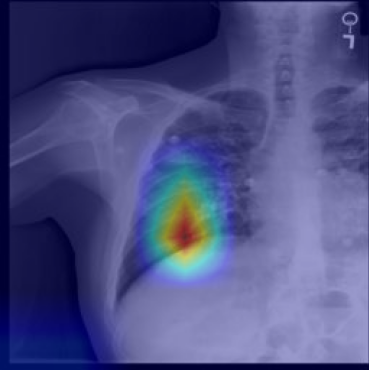}
        \parbox[c][3.5em][t]{\linewidth}{\centering \small Pneumonia}
	\end{minipage}
    \hspace{0.01\linewidth}
    \begin{minipage}{0.15\linewidth}
		\centering
		\includegraphics[width=\linewidth]{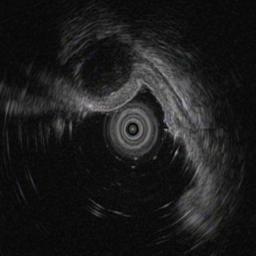}
        \includegraphics[width=\linewidth]{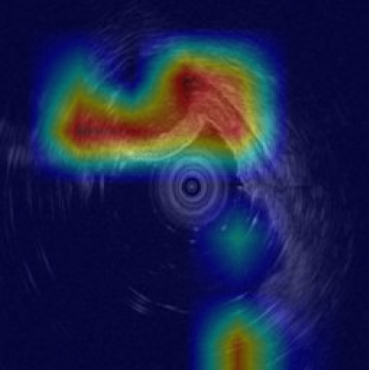}
        \parbox[c][3.5em][t]{\linewidth}{\centering \small low-echoic \\ mass}
	\end{minipage}
    \hspace{0.01\linewidth}
    \begin{minipage}{0.15\linewidth}
		\centering
		\includegraphics[width=\linewidth]{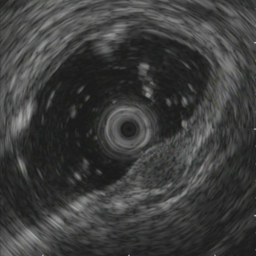}
        \includegraphics[width=\linewidth]{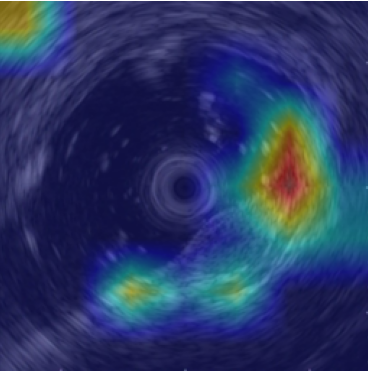}
        \parbox[c][3.5em][t]{\linewidth}{\centering \small protruded \\ into the \\ cavity}
	\end{minipage}
    \hspace{0.01\linewidth}
    \begin{minipage}{0.15\linewidth}
		\centering
		\includegraphics[width=\linewidth]{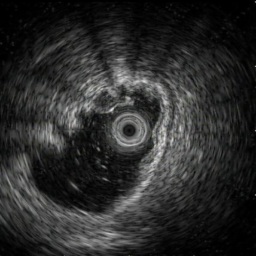}
        \includegraphics[width=\linewidth]{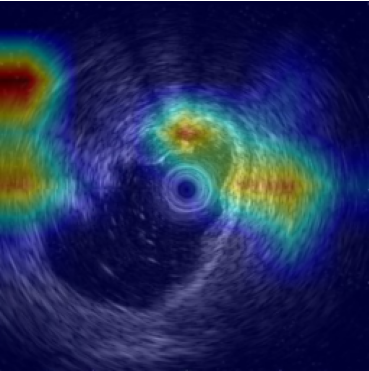}
        \parbox[c][3.5em][t]{\linewidth}{\centering \small clear \\ boundaries}
	\end{minipage}
    \caption{Visualization of learned token correspondence by our AGA. Highlighted pixels represent higher activation weights by corresponding word.}
    \label{fig:fig5}
\end{figure}

Fig.~\ref{fig:fig5} presents a qualitative visualization of the learned word-to-region correspondences facilitated by our AGA framework. The top row shows the original medical images, including both chest X-rays (CXR) and endoscopic ultrasound (EUS) images. The bottom row displays the corresponding heatmaps generated by our model, where warmer colors denote higher activation weights, indicating stronger associations between specific image regions and the given medical concepts. For Atelectasis and Pneumonia, the model focuses on appropriate pulmonary regions, demonstrating strong localization aligned with radiological pathology. For the SMTs domain, terms like low-echoic mass and protruded into the cavity activate precisely the relevant interior structures of the lesion in EUS images. The distinct and interpretable activation patterns validate the effectiveness of our AGA mechanism in achieving fine-grained multimodal alignment.

\subsection{Visualization of encoded image representations}

\begin{figure*}[htbp]
	\centering
    \begin{subfigure}[b]{0.32\textwidth}
        \centering
        \includegraphics[width=\textwidth]{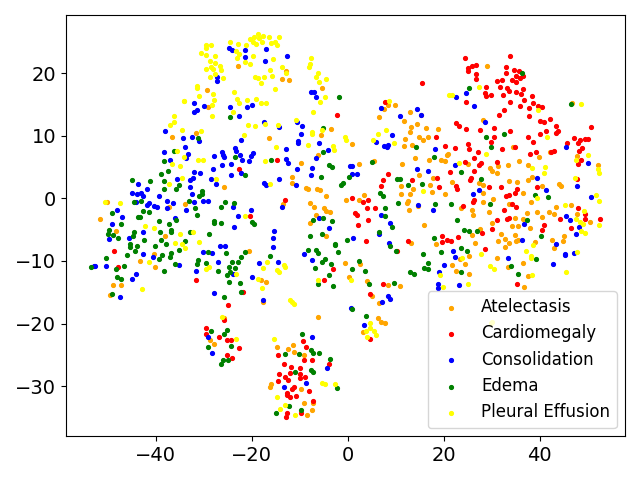}
        \caption{}
    \end{subfigure}
    \hfill
    \begin{subfigure}[b]{0.32\textwidth}
        \centering
        \includegraphics[width=\textwidth]{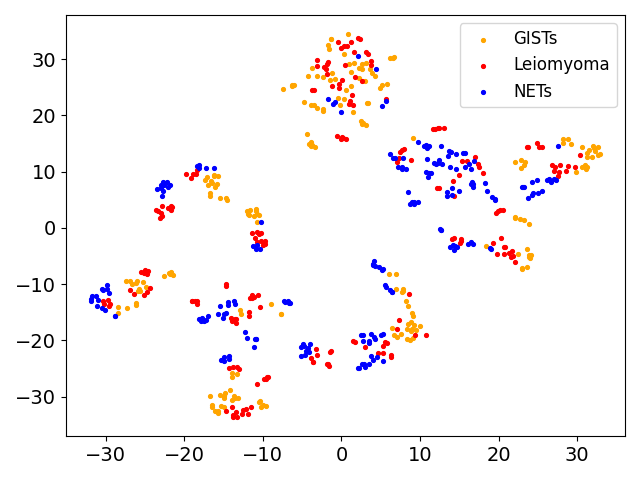}
        \caption{}
    \end{subfigure}
    \hfill
    \begin{subfigure}[b]{0.32\textwidth}
        \includegraphics[width=\textwidth]{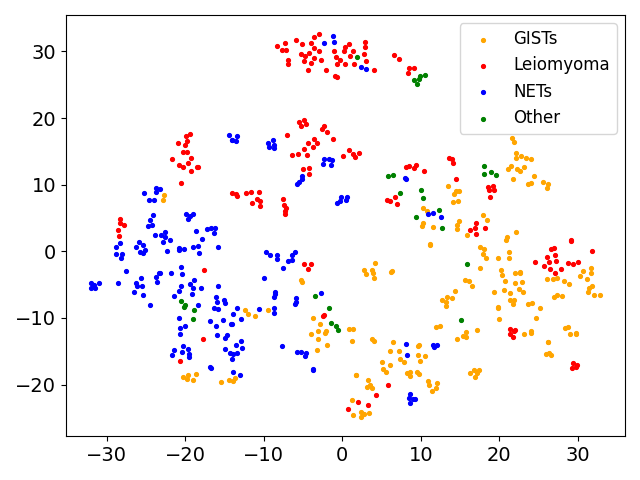}
        \caption{}
    \end{subfigure}
    \caption{. T-SNE visualizations of encoded image representations. Colors represent the ground truth disease types. Subfigures (a), (b), and (c) correspond to the results on the CheXpert 5×200, SMTs 3×200, and SMTs SN datasets, respectively.}
    \label{fig:fig6}
\end{figure*}

To qualitatively assess the effectiveness of our AGA framework, we visualize the learned image representations using t-SNE on three datasets: CheXpert 5×200, SMTs 3×200, and SMTs SN. The results are shown in Fig.~\ref{fig:fig6}. In subfigure (a), despite the limited supervision and complex semantics of chest X-ray images, our model captures meaningful intra-class patterns, with moderate separation between disease categories such as Atelectasis, Cardiomegaly, and Pleural Effusion. Some overlaps are observed between semantically similar conditions, which may stem from the ambiguity in clinical labels. In contrast, subfigure (b) shows more distinct and compact clusters on the SMTs 3×200 dataset, suggesting that the structured and domain-specific textual annotations in our private dataset enhance the semantic alignment between visual and textual modalities. Subfigure (c) further demonstrates the generalization ability of our method on SMTs SN, a more challenging dataset with a broader category distribution. The representation space remains clearly structured, and additional disease types such as "Other" are accommodated without disrupting the separability of core classes like GISTs, Leiomyoma, and NETs. These results confirm that our grouping-based strategy enabled by the IGA loss and BCGA module, effectively promotes intra-class cohesion and inter-class discrimination. The visualizations provide strong qualitative evidence that AGA learns semantically grounded and context-consistent representations, which serve as a robust foundation for downstream tasks such as classification and retrieval.


\section{Conclusion}

In this work, we propose an AGA framework for cross-modal medical visual representation learning. Our approach introduces an IGA loss that captures fine-grained associations between textual tokens and their corresponding group representations, supported by a BCGA module to refine group interactions and promote semantic consistency. To address the granularity inconsistency across datasets and enable adaptive grouping, we introduce a dynamic grouping threshold gate that learns to adjust thresholds during training, facilitating more flexible alignment based on the data. We conduct extensive evaluations across image-to-text retrieval, supervised classification, and zero-shot classification tasks on both public and private datasets. The results demonstrate that our model outperforms existing methods, especially in low-data regimes. Visualization analyses including feature scatter plots and activation heatmaps further confirm that the learned representations exhibit strong intra-class cohesion and precise semantic grounding. Compared with MIMIC-CXR, the SMTs datasets exhibit more structured descriptions, leading to nearly identical visual and textual grouping thresholds and stronger word-region correlations. 

Since our work primarily focuses on medical visual representation learning, we did not evaluate performance on text-based downstream tasks, which can be regarded as a limitation of this study. In future work, we plan to extend the grouping strategy to the sample level to enable alignment across groups of samples. We also intend to integrate our approach with generation-based pre-training methods to facilitate joint learning of image and textual features.

\section*{Declaration of Interests}
Authors declare that they have no conflict of interest.

\section*{Acknowledgments}
This work is partially supported by National Natural Science Foundation of China (62376231), Sichuan Science and Technology Program (24NSFSC1070), Fundamental Research Funds for the Central Universities (2682025ZTPY052, 2682023ZDPY001).

\bibliographystyle{unsrtnat}
\bibliography{references}  






\end{document}